\documentclass{article}
\usepackage{spconf,amsmath,graphicx}
\usepackage{amsmath,amssymb,amsfonts}
\usepackage{booktabs}
\usepackage{multirow}
\usepackage[table,xcdraw]{xcolor}
\usepackage[ruled,vlined, linesnumbered, lined,boxed,commentsnumbered]{algorithm2e}
\usepackage{subcaption}

\title{Multiple Object Tracking based on Occlusion-Aware Data Association}
\title{Multiple Object Tracking based on Occlusion-Aware Embedding Consistency Learning}
\name{Yaoqi Hu, Axi Niu, Yu Zhu, Qingsen Yan, Jinqiu Sun*, Yanning Zhang \thanks{This work was funded in part by the Project of the National Natural Science Foundation of China under Grant 61871328, Natural Science Basic Research Program of Shaanxi under Grant 2021JCW-03, as well as the Joint Funds of the National Natural Science Foundation of China under Grant U19B2037.).
(*Corresponding author: Jinqiu Sun.)}}

\address{Northwestern Polytechnical University}
\begin{document}
%
\maketitle
\begin{abstract}
The Joint Detection and Embedding (JDE) framework has achieved remarkable progress for multiple object tracking. Existing methods often employ extracted embeddings to re-establish associations between new detections and previously disrupted tracks. However, the reliability of embeddings diminishes when the region of the occluded object frequently contains adjacent objects or clutters, especially in scenarios with severe occlusion. To alleviate this problem, we propose a novel multiple object tracking method based on visual embedding consistency, mainly including: 1) \textbf{O}cclusion \textbf{P}rediction \textbf{M}odule (\textbf{OPM}) and 2) \textbf{O}cclusion-\textbf{A}ware \textbf{A}ssociation \textbf{M}odule (\textbf{OAAM}). The \textbf{OPM} predicts occlusion information for each true detection, facilitating the selection of valid samples for consistency learning of the track's visual embedding.
The \textbf{OAAM} leverages occlusion cues and visual embeddings to generate two separate embeddings for each track, guaranteeing consistency in both unoccluded and occluded detections. By integrating these two modules, our method is capable of addressing track interruptions caused by occlusion in online tracking scenarios. Extensive experimental results demonstrate that our approach achieves promising performance levels in both unoccluded and occluded tracking scenarios.

\end{abstract}

\vspace{-10pt}
\section{Introduction}
\vspace{-6pt}
\label{sec:intro}
Online \textbf{M}ultiple \textbf{O}bject \textbf{T}racking (\textbf{MOT}) plays an important role in video surveillance~\cite{oh2011large}, autonomous driving~\cite{sun2020scalability}, and many other fields~\cite{zhang2020multi}. Given the input video frames, online MOT methods locate track positions frame by frame while giving unique identifications for each of them. Though online MOT methods have made significant progress in recent decades, the persistence of tracking failures due to occlusion remains a major obstacle to the widespread practical adoption of such methods. In order to resolve the above problem, the common practice involves a bottom-up association strategy~\cite{Zhang_Sheng_Wu_Wang_Lyu_Ke_Xiong_2020, du2023strongsort}, which prioritizes track-detection association as the initial step, followed by track-track association. Within the track-track association phase, visual embeddings are extracted for each track using an external re-identification (re-ID) network, which is then employed to compute the similarity between different tracks. Although those methods can recover the loss track even if some detections are missed for a long range due to occlusion, the re-identification network's training and inference procedures are time-consuming~\cite{ Liu_Li_Bai_Wang_Wang_2021, Liu_Chu_Liu_Yu_2020, aharon2022bot}. Furthermore, the adopted association strategy necessitates the use of current data to amend the historical segment of the track, which is not suitable for online tracking. 

To mitigate the above issues, methods based on \textbf{JDE} paradigm are becoming prevalent, which learn both detection and re-ID embedding extraction in a shared model~\cite{wang2020towards}. 
In particular, FairMOT~\cite{zhang_wang_wang_zeng_liu_2021} achieves a good balance between tracking performance and efficiency by adopting a point-based framework. It follows MOTDT~\cite{Chen_Ai_Zhuang_Shang_2018} and uses a hierarchical track-detection association method, which combines Kalman filter, re-ID embeddings, and Intersection-over-Union (IOU) metrics to facilitate the matching of tracks with detections. 
 OUTrack~\cite{liu_chen_chu_yuan_liu_zhang_yu_2022} optimizes the visual embedding by the unsupervised reID learning which does not require any (pseudo) identity information nor suffer from the scalability issue. What's more, it designs an occlusion estimation module which is used to recover the miss detections caused by occlusions. Although these methods have made notable progress, they depend on the assumption that the importance of visual embedding is higher than motion information, which is not applicable with severe occlusion. Moreover, these methods do not differentiate between the visual embeddings of occluded and unoccluded detections when updating the track's embeddings.

To address these issues, we propose a novel multiple object tracking method based on visual embedding consistency to conduct track-detection associations in severe occlusions. The consistency means the visual embeddings of all detections belonging to the same track should be similar. During the track-detection association process, a \textbf{S}mooth \textbf{V}isual \textbf{E}mbedding (\textbf{SVE}) is maintained to characterize the visual feature for each track. The \textbf{SVE} is iteratively refined and updated by the current matched detection. When the detection is occluded, the \textbf{SVE} is updated with the embeddings of occluding detections. This update disrupts the consistency of the visual embedding in the subsequent association process. In this paper, we decompose the \textbf{SVE} into two embeddings: long-term embedding and short-term embedding. Long-term embedding characterizes the visual feature for unoccluded detections in a track. Short-term embedding characterizes the visual feature for occluded detections in a track. The two embeddings can realize the consistency of the track's visual embedding in both unoccluded and occluded situations.

Specifically,  we devise two modules: \textbf{O}cclusion \textbf{P}rediction \textbf{M}odule (\textbf{OPM}) and \textbf{O}cclusion-\textbf{A}ware \textbf{A}ssociation \textbf{M}odule (\textbf{OAAM}). The \textbf{OPM} predicts occlusion information for each detection, indicating the degree a detection's position is occluded. Through the information, we can select unoccluded true detections as valid samples to conduct the consistency learning of the track's visual embedding. In the \textbf{OAAM}, we design a two-stage track-detection association strategy. 
In each stage, we get the matching result by computing the similarities between tracks and detections based on their visual embeddings and bounding boxes. In stage one, the short-term embedding is treated as the track's visual embedding and updated by each matched detection's visual embedding. In stage two, the long-term embedding is treated as the track's visual embedding and only updated by the matched unoccluded detection's visual embedding. The two-stage strategy dynamically associates and updates the tracks for unocclued and occluded detections.

To summarize, we make the following contributions: 
\begin{itemize}

\item We conceptualize the track-detection association under severe occlusion as the task of preserving visual embedding consistency for both unoccluded and occluded detections, greatly simplifying the association process. 

\item We design long-term embedding and short-term embedding to characterize unoccluded and occluded detections, which help maintain the consistency of the track’s visual embedding in severe occluded situations. 

\item Based on the above embeddings, we build \textbf{O}cclusion \textbf{P}rediction \textbf{M}odule and \textbf{O}cclusion-\textbf{A}ware \textbf{A}ssociation \textbf{M}odule to dynamically associate and update the tracks for unoccluded and
occluded detections to improve the matching result under severe occlusion.



\end{itemize}

\vspace{-10pt}
\begin{figure*}
\centering
\includegraphics[width=1.0\textwidth]{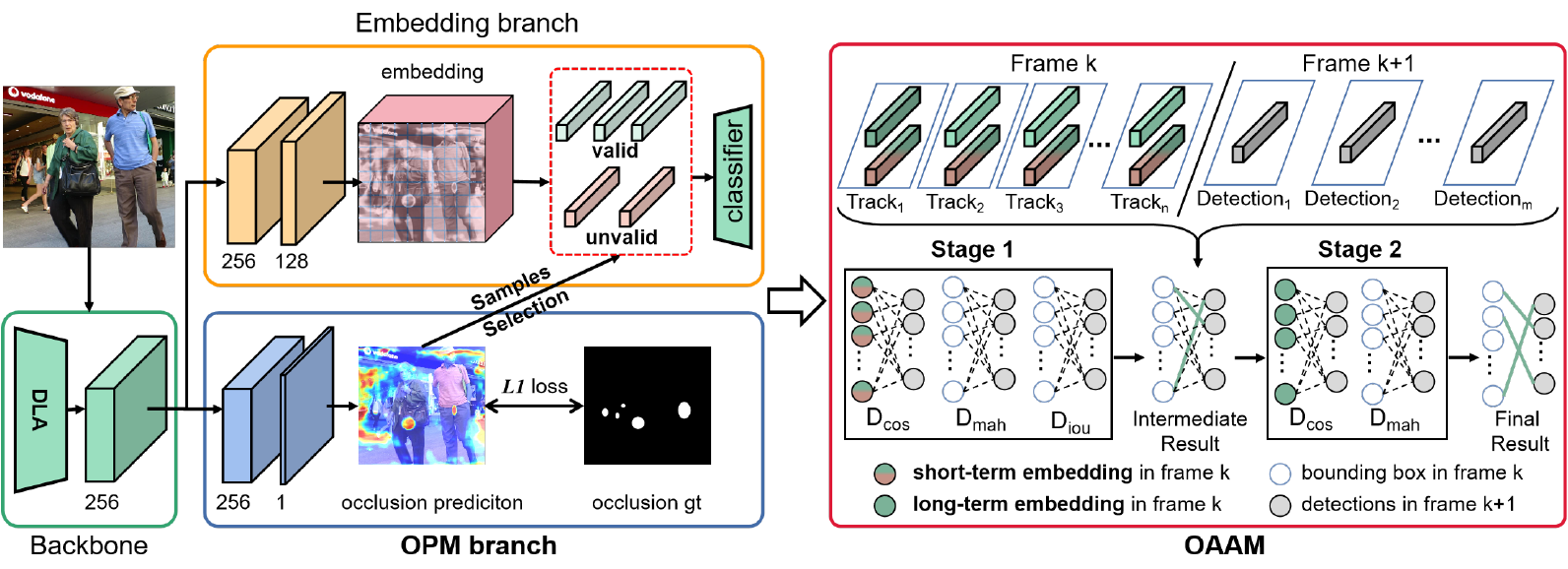}
\caption{Our method mainly includes: 1) \textbf{O}cclusion \textbf{P}rediction \textbf{M}odule (\textbf{OPM}) and 2) \textbf{O}cclusion-\textbf{A}ware \textbf{A}ssociation \textbf{M}odule (\textbf{OAAM}). The \textbf{OPM} produces an occlusion prediction map aiding the sample selection in Embedding Learning. The \textbf{OAAM} is used to build two embeddings based on occlusion prediction to convey the matching between tracks and detections. Better view in color.}
\vspace{-12pt}
\label{fig:framework}
\end{figure*}

\vspace{-4pt}
\section{Proposed Method}
\vspace{-4pt}
\label{sec:method}

This section begins with the introduction of our method. As shown in Fig.~\ref{fig:framework}, our method mainly consists of \textbf{O}cclusion \textbf{P}rediction \textbf{M}odule (\textbf{OPM}) and \textbf{O}cclusion-\textbf{A}ware \textbf{A}ssociation \textbf{M}odule (\textbf{OAAM}). Next, we will describe them in detail.


\vspace{-4pt}
\subsection{Occlusion Prediction Module}
The OPM is used to generate the occlusion information for each detection. Inspired by FairMOT~\cite{zhang_wang_wang_zeng_liu_2021}, we use the CenterNet~\cite{zhou2019objects} as our backbone. By adding an extra OPM branch in the CenterNet architecture, we can predict the degree of occlusion for each detection. 

\textbf{Structure of OPM.} Our OPM branch has a similar structure to the existing branches in CenterNet. Specifically, the feature $\textbf{F}$ of $256\times{H} \times{W}$ extracted by the backbone is input into our OPM branch. Then, it is passed through one $3\times{3}$ convolutional layer with 256 channels and one $1\times{1}$ convolution with 1 channel to produce the desired output. The feature map produced by our OPM branch is denoted as the occlusion prediction map $\hat{\textbf{S}}$. Its dimension is $1\times{H} \times{W}$. The response values at each location of the occlusion prediction map are bounded within the range of 0 to 1. A response value nearing 1 indicates that the detection associated with that particular position is experiencing minimal occlusion, while a response value of 0 signifies that the detection is fully occluded.

\textbf{Training of OPM.} For each Ground Truth bounding box $\textbf{b}^i=(x_1^i, y_1^i, x_2^i, y_2^i, v^i)$, 
includes the left, top, right, and bottom coordinates and visibility, we compute the detection center $(c_x^i, c_y^i)$ as $c_x^i=\frac{x_1^i + x_2^i}{2}$ and $c_y^i=\frac{y_1^i + y_2^i}{2}$. After mapping the detection center onto the occlusion prediction map, we obtain the corresponding position as $(\hat{c}_x^i, \hat{c}_y^i) = (\lfloor\frac{c_x^i}{4}\rfloor, \lfloor\frac{c_y^i}{4}\rfloor)$. We define $s^i = 1-v^i$ as the ground truth of the occluded degree for the detection $i$. Then the response of GT occlusion map $\textbf{S}$ at the location $(x, y)$ is computed as $S_{xy} = \sum_i^N{\delta{(x-\hat{c}_x^i,y-\hat{c}_y^i)}}$where $N$ represents the number of detections in the image and $\delta$ is the two-dimensional impulse function.
Then, we use the L1 loss to train our OPM branch, which considers only the prediction results at the position of the truth detection center:
\begin{equation}
    L_{occ}=\frac{1}{N}\sum_{xy}{
    \left\{
    \begin{aligned}
        &\vert S_{x,y} - \hat{S}_{x,y}\vert && S_{x,y}>0 \\
        &0 && S_{x,y}=0 
    \end{aligned}
    \right.
    }.
\end{equation}

\textbf{Training of Embedding branch.} The output map of embedding branch is $\textbf{E}\in{R^{128\times{H}\times{W}}}$. We take $\textbf{E}_{(\hat{c}_x^i, \hat{c}_y^i)}\in{R^{128}}$ as the visual embedding of the detection centered at $(\hat{c}_x^i, \hat{c}_y^i)$. These features are learned by a multiclass classifier, where the labels are the identity numbers of the corresponding detections. In contrast to the training process for embedding branches in FairMOT, our method involves a sample selection operation. We select unoccluded detections as valid samples with occlusion information provided by OPM. We use valid samples to train our embedding branch, which helps to  conduct the consistency
learning of the track’s visual embedding.

\vspace{-8pt}
\subsection{Occlusion-Aware Association Module}
Here, we first design long-term embedding and short-term embedding to characterize the visual feature for unoccluded and occluded detections in the track. In OAAM, we design a two-stage track-detection association strategy by applying the two embeddings into different stages. 

\textbf{Two visual embeddings.} For each track, we define two visual embeddings: long-term embedding $\textbf{F}^l_i$ and short-term embedding $\textbf{F}^s_i$. We divide detections into two groups based on their degree of being occluded: occluded detections and non-occluded detections. The long-term embedding is updated through the visual embedding from matched unoccluded detections. Conversely, the short-term embedding is updated using the visual embedding of every matched detection without taking into consideration their degree of occlusion. For each track, we define the updated function as below:
\begin{equation}
    \textbf{F}^l_i = norm(\textbf{F}^l_i + \alpha*\mathbb{I}(\hat{\textbf{S}}_{(\hat{c}_x^i, \hat{c}_y^i)} > 0)*norm(\textbf{E}_{(\hat{c}_x^i, \hat{c}_y^i)}^t)),
    \label{eq:long-term}
\end{equation}

\vspace{-12pt}
\begin{equation}
    \textbf{F}^s_i = norm(\textbf{F}^s_i + \beta_i^t*norm(\textbf{E}_{(\hat{c}_x^i, \hat{c}_y^i)}^t)),
    \label{eq:short-term}
\end{equation}
where $\alpha$ is a fixed weight, usually taking the value of 0.2, and $\beta$ is used to dynamically adjust the weight of the detection according to the degree of its occlusion.

\begin{figure}
\centering
\includegraphics[width=0.45\textwidth]{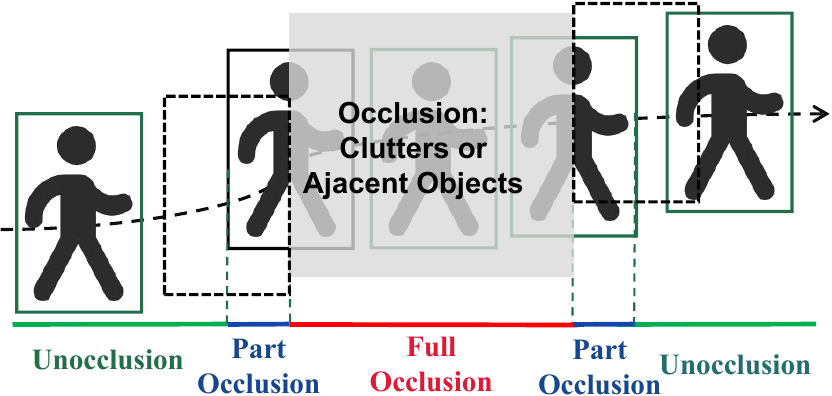}
\caption{Detections have three states: 1) Unocclusion, 2) Part Occlusion, 3) and Full Occlusion. Part Occlusion means the occluded detection is still detected. }
\vspace{-6pt}
\label{fig:process_occl}
\end{figure}

\begin{figure}[t!]
    \centering
    \begin{subfigure}[t]{0.23\textwidth}
           \centering
           \includegraphics[width=0.85\textwidth]{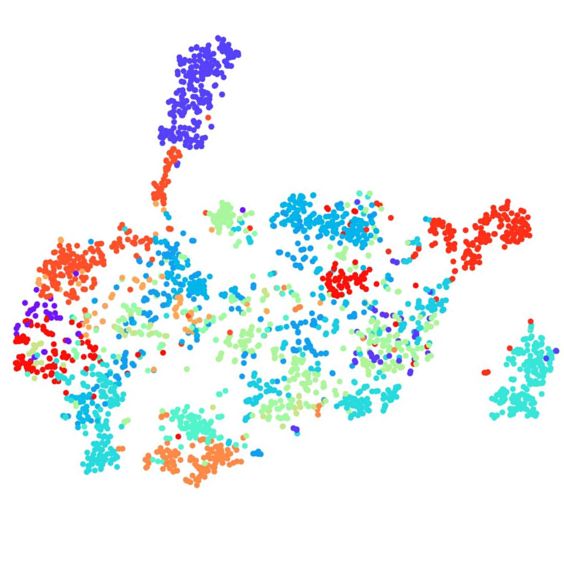}
           \vspace{-8pt}
            \caption{FairMOT}
            \vspace{-6pt}
            \label{fig:a}
    \end{subfigure}
    \begin{subfigure}[t]{0.23\textwidth}
            \centering
    \includegraphics[width=0.85\textwidth]{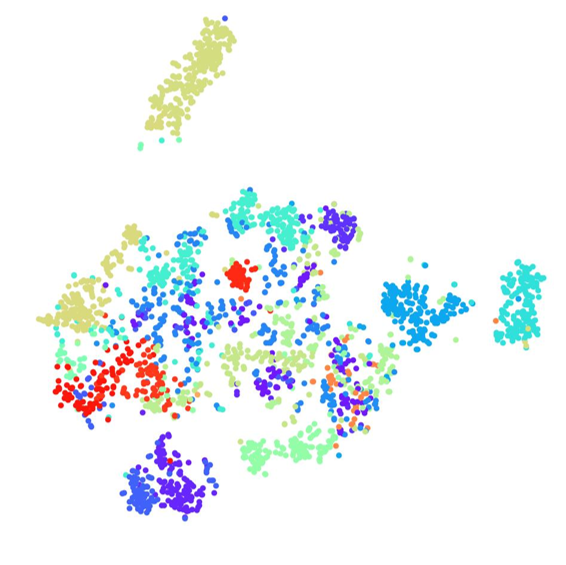}
            \vspace{-8pt}
            \caption{Ours}
            \vspace{-6pt}
            \label{fig:b}
    \end{subfigure}
    \caption{Visualization of learned embeddings for unoccluded detections by t-SNE~\cite{van2008visualizing}. (a): the compared method (b): the proposed method. Different colors indicate different tracks.}
    \label{fig:occl_state}
\end{figure}

\textbf{Track-detection Association.} In different frames, a detection may have three states: 1) non-occlusion, 2) part-occlusion, 3) and full-occlusion, as shown in Fig.~\ref{fig:occl_state}. The whole process of the occlusion is shown in Fig.~\ref{fig:process_occl}. In our track-detection association algorithm, we design a two-stage strategy to deal with different states. 
Before the association step, we employ a Kalman filter to forecast the position and size of each track's bounding box in the current frame. 

In stage one, we compute three distances to address the matching problem arisen during the transition from non-occlusion to partial occlusion. Specifically, we compute the Mahalanobis distance $D_{mah}$ and Intersection over Union (IOU) $D_{iou}$ between predicted bounding boxes of tracks and current detections, following SORT~\cite{wojke2017simple}. We compute the cosine distance $D_{cos}$ based on short-term embedding. Then, we fuse them into a single similarity metric by weighting parameters: $D = \gamma*(\lambda*D_{cos} + (1 - \lambda)*D_{mah}) + (1-\gamma)*(1 - D_{iou})$, where $\gamma$ and $\lambda$ are set as 0.8 and 0.9 respectively following~\cite{zhang_wang_wang_zeng_liu_2021}.

In stage two, we compute two distances to address the matching problem that arises during the transition from partial occlusion to non-occlusion. Specifically, for unmatched tracks and detections in the previous stage,  we compute the Mahalanobis distance $D_{mah}$ between predicted bounding boxes of tracks and current detections. We compute cosine distance $D_{cos}$ between long-term embeddings of unmatched tracks and visual embeddings of detections.  After the track-detection association, we update the long-term and short-term embedding of all matched tracks by Eq.~\ref{eq:long-term} and \ref{eq:short-term}. We initialize unmatched detections into new tracks. For unmatched tracks, we keep up to 30 frames, spanning the time from full occlusion to reappearance.


\vspace{-10pt}
\section{Experiments}
\vspace{-4pt}

\subsection{Implementation Details}
\vspace{-4pt}
We evaluate our method on multiple MOT datasets like MOT17~\cite{milan2016mot16} and MOT20~\cite{dendorfer2020mot20}. Following the training settings in FairMOT~\cite{zhang_wang_wang_zeng_liu_2021}, we pre-train our model on CrowdHuman~\cite{shao2018crowdhuman} for 60 epochs with a batch size of 40, the initial learning rate of 1.83e-4 under decay of 0.1 for 50 epochs. For testing, we finetune the pre-trained model on MOT17 and MOT20 separately for 30 epochs with a batch size of 40, the initial learning rate of 1.83e-4 under decay of 0.1 for 20 epochs. For more details, please refer FairMOT~\cite{zhang_wang_wang_zeng_liu_2021}. We use HOTA~\cite{luiten2021hota} as the dominant evaluation metric. HOTA is a sophisticated combination of the detection score and the association score, which maintains a balance between the accuracy of detection and association. We also use the metrics MOTA, IDs, IDF1~\cite{ristani2016performance}, FP, and FN to evaluate tracking performance.

\begin{table}[t]
\centering
\caption{Comparison of different Data Association Methods.}
\vspace{-6pt}
\resizebox{0.4\textwidth}{!}{%
\begin{tabular}{c|c|c|c|c|c}
\toprule
\hline
\textbf{Method} & \textbf{IDF1$\uparrow$} & \textbf{MOTA$\uparrow$} & \textbf{FP$\downarrow$} & \textbf{FN$\downarrow$} & \textbf{IDs$\downarrow$} \\ \midrule
FairMot & 70.8 & \textbf{67.7} & 3220 & 13878 & 421 \\
Byte & 71.1 & 67.8 & 3696 & \textbf{13291} & 301 \\
Ours & \textbf{73.5(+2.7)} & \textbf{67.5(-0.2)} & \textbf{2045} & 15243 & \textbf{362} \\
\hline
\bottomrule
\end{tabular}%
}
\label{tab:Association}
\end{table}

\vspace{-10pt}
\subsection{Ablation Studies}
Here, we perform ablation studies to illustrate the significance of the designed modules in our method. In this section, our analysis is performed on MOT17, half of which is used for training and half for testing.

\textbf{Visual Embedding Consistency.}~Different from the training way in FairMOT, we select unoccluded objects as training samples to learn the embedding consistency for the track.  We employ t-SNE to visualize the learned embeddings of unoccluded objects from our method and FairMOT separately. As shown in Fig.~\ref{fig:occl_state}, for the same object, the learned embeddings from our method are better grouped than those from FairMOT. It means the extracted embedding learned by the embedding branch trained with selected samples performs better for maintaining the visual embedding consistency.

\textbf{Comparison of different Association Methods.}~We compare our \textbf{O}cclusion-\textbf{A}ware \textbf{A}ssociation \textbf{M}odule (\textbf{OAAM}) with association strategies in FairMOT~\cite{zhang_wang_wang_zeng_liu_2021}, Bytetrack~\cite{zhang2022bytetrack}. 
As shown in Tab.~\ref{tab:Association}, our \textbf{OAAM} superiors other strategies with higher MOTA, IDF1, and lower IDs. Our \textbf{OAAM} dynamically utilizes the embedding consistency for unoccluded detections to re-establish associations between newly detected objects and previously disrupted tracks.

\vspace{-10pt}
\subsection{Benchmark Results}
\vspace{-6pt}
We compare our method with the state-of-the-art trackers on the test set of MOT17 and MOT20 under the private detection protocol in Tab.~\ref{tab:mot17} and Tab.~\ref{tab:mot20}. All results are obtained from the published codes or corresponding papers. For MOT17, our method achieves the best IDF1 and HOTA. This observation demonstrates that our training strategy effectively preserves the consistency of visual embeddings for unoccluded detections, a crucial factor in re-establishing connections between previously interrupted tracks and current detections. Our MOTA metric is 0.7 lower than the best 75.2. This occurs because, in some cases, abrupt changes from occluders result in a significant alteration to the occluded detection's visual embedding, leading to the failure of track-detection associations. In our future work, we will try to solve this problem. For MOT20, wo also obtains similar achievement to MOT17 with the best MOTA, IDF1, and HOTA metrics.

\begin{table}[t]
\centering
\caption{State-of-the-art methods on the test sets of MOT17.}
\vspace{-6pt}
\resizebox{0.5\textwidth}{!}{%
\begin{tabular}{c|ccccc}
\toprule
\hline
\textbf{Tracker} & \textbf{MOTA$\uparrow$} & \textbf{IDF1$\uparrow$} & \textbf{HOTA$\uparrow$} & \textbf{FP$\downarrow$} & \textbf{FN$\downarrow$}  \\ \midrule
TransCenter~\cite{xu2021transcenter} & 73.2 &62.2 &54.5 &\textbf{23112} &123738  \\
FairMOT~\cite{zhang_wang_wang_zeng_liu_2021} & 73.7 &72.3 &59.3 &27507 &117477  \\
PermaTrackPr~\cite{tokmakov2021learning} & 73.8 &68.9 &55.5 &28998 &115104  \\
CSTrack~\cite{liang2022rethinking} & 74.9 &72.6 &59.3 &23847 &114303  \\
TransTrack~\cite{chu2023transmot} & \textbf{75.2} &63.5 &54.1 &50157 &\textbf{86442}  \\
Ours & 74.5 & \textbf{74.5} & \textbf{60.8} & 29805  & 111897  \\ 


\hline
\bottomrule
\end{tabular}%
}
\label{tab:mot17}
\end{table}

  \begin{table}[t]
\centering
\caption{State-of-the-art methods on the test sets of MOT20.}
\vspace{-6pt}
\resizebox{0.5\textwidth}{!}{%
\begin{tabular}{c|ccccc}
\toprule
\hline
\textbf{Tracker} & \textbf{MOTA$\uparrow$} & \textbf{IDF1$\uparrow$}  & \textbf{HOTA$\uparrow$}  & \textbf{FP$\downarrow$} & \textbf{FN$\downarrow$} \\ 
\midrule
FairMOT~\cite{zhang_wang_wang_zeng_liu_2021} &61.8 &67.3 &54.6 &103440 &88901 \\
TransCenter~\cite{xu2021transcenter} &61.9 &50.4 &- &45895 &146347  \\
TransTrack~\cite{chu2023transmot} &65.0 &59.4 &48.5 &27197 &150197  \\
CSTrack~\cite{liang2022rethinking} &66.6 &68.6 &54.0 &\textbf{25404} &144358 \\
Ours &\textbf{68.1} & \textbf{70.3}&\textbf{ 57.8}& 26536& \textbf{129045}\\
\hline
\bottomrule
\end{tabular}%
}
\label{tab:mot20}
\end{table}

\vspace{-10pt}
\section{Conclusion}
\vspace{-6pt}
In this paper, we propose a novel multiple object tracking method based on visual embedding consistency to conduct online data association in severe occlusions. We design long-term embedding and short-term embedding to characterize the visual feature for unoccluded and occluded detections in the track, which maintain the consistency of the track’s visual embedding in severe occluded situations. Then, we adopt a two-stage track-detection association strategy to dynamically associate and update the tracks for unoccluded and occluded detections by applying the two embeddings into different stages. Extensive experiments have demonstrated the effectiveness of our methods. In the future, we will further try to pursue better visual embedding consistency for occluded situations with other degradations.

\newpage
\bibliographystyle{IEEEbib}

\bibliography{strings,refs}

\begin{thebibliography}{10}

\bibitem{oh2011large}
Sangmin Oh, Anthony Hoogs, Amitha Perera, Naresh Cuntoor, Chia-Chih Chen,
  Jong~Taek Lee, Saurajit Mukherjee, JK~Aggarwal, Hyungtae Lee, Larry Davis,
  et~al.,
\newblock ``A large-scale benchmark dataset for event recognition in
  surveillance video,''
\newblock in {\em CVPR}, 2011.

\bibitem{sun2020scalability}
Pei Sun, Henrik Kretzschmar, Xerxes Dotiwalla, Aurelien Chouard, Vijaysai
  Patnaik, Paul Tsui, James Guo, Yin Zhou, Yuning Chai, Benjamin Caine, et~al.,
\newblock ``Scalability in perception for autonomous driving: Waymo open
  dataset,''
\newblock in {\em CVPR}, 2020.

\bibitem{zhang2020multi}
Ruiheng Zhang, Lingxiang Wu, Yukun Yang, Wanneng Wu, Yueqiang Chen, and Min Xu,
\newblock ``Multi-camera multi-player tracking with deep player identification
  in sports video,''
\newblock {\em PR}, 2020.

\bibitem{Zhang_Sheng_Wu_Wang_Lyu_Ke_Xiong_2020}
Yang Zhang, Hao Sheng, Yubin Wu, Shuai Wang, Weifeng Lyu, Wei Ke, and Zhang
  Xiong,
\newblock ``Long-term tracking with deep tracklet association,''
\newblock 2020.

\bibitem{du2023strongsort}
Yunhao Du, Zhicheng Zhao, Yang Song, Yanyun Zhao, Fei Su, Tao Gong, and
  Hongying Meng,
\newblock ``Strongsort: Make deepsort great again,''
\newblock {\em TMM}, 2023.

\bibitem{Liu_Li_Bai_Wang_Wang_2021}
Yating Liu, Xuesong Li, Tianxiang Bai, Kunfeng Wang, and Fei-Yue Wang,
\newblock ``Multi-object tracking with hard-soft attention network and
  group-based cost minimization,''
\newblock {\em Neurocomputing}, 2021.

\bibitem{Liu_Chu_Liu_Yu_2020}
Qiankun Liu, Qi~Chu, Bin Liu, and Nenghai Yu,
\newblock ``Gsm: Graph similarity model for multi-object tracking.,''
\newblock in {\em IJCAI}, 2020.

\bibitem{aharon2022bot}
Nir Aharon, Roy Orfaig, and Ben-Zion Bobrovsky,
\newblock ``Bot-sort: Robust associations multi-pedestrian tracking,''
\newblock {\em arXiv:2206.14651}, 2022.

\bibitem{wang2020towards}
Zhongdao Wang, Liang Zheng, Yixuan Liu, Yali Li, and Shengjin Wang,
\newblock ``Towards real-time multi-object tracking,''
\newblock in {\em ECCV}, 2020.

\bibitem{zhang_wang_wang_zeng_liu_2021}
Yifu Zhang, Chunyu Wang, Xinggang Wang, Wenjun Zeng, and Wenyu Liu,
\newblock ``Fairmot: On the fairness of detection and re-identification in
  multiple object tracking,''
\newblock {\em IJCV}, 2021.

\bibitem{Chen_Ai_Zhuang_Shang_2018}
Long Chen, Haizhou Ai, Zijie Zhuang, and Chong Shang,
\newblock ``Real-time multiple people tracking with deeply learned candidate
  selection and person re-identification,''
\newblock in {\em ICME}, 2018.

\bibitem{liu_chen_chu_yuan_liu_zhang_yu_2022}
Qiankun Liu, Dongdong Chen, Qi~Chu, Lu~Yuan, Bin Liu, Lei Zhang, and Nenghai
  Yu,
\newblock ``Online multi-object tracking with unsupervised re-identification
  learning and occlusion estimation,''
\newblock {\em Neurocomputing}, 2022.

\bibitem{zhou2019objects}
Xingyi Zhou, Dequan Wang, and Philipp Kr{\"a}henb{\"u}hl,
\newblock ``Objects as points,''
\newblock {\em arXiv:1904.07850}, 2019.

\bibitem{wojke2017simple}
Nicolai Wojke, Alex Bewley, and Dietrich Paulus,
\newblock ``Simple online and realtime tracking with a deep association
  metric,''
\newblock in {\em ICIP}, 2017.

\bibitem{van2008visualizing}
Laurens Van~der Maaten and Geoffrey Hinton,
\newblock ``Visualizing data using t-sne.,''
\newblock {\em JMLR}, 2008.

\bibitem{milan2016mot16}
Anton Milan, Laura Leal-Taix{\'e}, Ian Reid, Stefan Roth, and Konrad Schindler,
\newblock ``Mot16: A benchmark for multi-object tracking,''
\newblock {\em arXiv:1603.00831}, 2016.

\bibitem{dendorfer2020mot20}
Patrick Dendorfer, Hamid Rezatofighi, Anton Milan, Javen Shi, Daniel Cremers,
  Ian Reid, Stefan Roth, Konrad Schindler, and Laura Leal-Taix{\'e},
\newblock ``Mot20: A benchmark for multi object tracking in crowded scenes,''
\newblock {\em arXiv:2003.09003}, 2020.

\bibitem{shao2018crowdhuman}
Shuai Shao, Zijian Zhao, Boxun Li, Tete Xiao, Gang Yu, Xiangyu Zhang, and Jian
  Sun,
\newblock ``Crowdhuman: A benchmark for detecting human in a crowd,''
\newblock {\em arXiv:1805.00123}, 2018.

\bibitem{luiten2021hota}
Jonathon Luiten, Aljosa Osep, Patrick Dendorfer, Philip Torr, Andreas Geiger,
  Laura Leal-Taix{\'e}, and Bastian Leibe,
\newblock ``Hota: A higher order metric for evaluating multi-object tracking,''
\newblock {\em IJCV}, 2021.

\bibitem{ristani2016performance}
Ergys Ristani, Francesco Solera, Roger Zou, Rita Cucchiara, and Carlo Tomasi,
\newblock ``Performance measures and a data set for multi-target, multi-camera
  tracking,''
\newblock in {\em ECCV}, 2016.

\bibitem{zhang2022bytetrack}
Yifu Zhang, Peize Sun, Yi~Jiang, Dongdong Yu, Fucheng Weng, Zehuan Yuan, Ping
  Luo, Wenyu Liu, and Xinggang Wang,
\newblock ``Bytetrack: Multi-object tracking by associating every detection
  box,''
\newblock in {\em ECCV}, 2022.

\bibitem{xu2021transcenter}
Yihong Xu, Yutong Ban, Guillaume Delorme, Chuang Gan, Daniela Rus, and Xavier
  Alameda-Pineda,
\newblock ``Transcenter: Transformers with dense queries for multiple-object
  tracking,''
\newblock 2021.

\bibitem{tokmakov2021learning}
Pavel Tokmakov, Jie Li, Wolfram Burgard, and Adrien Gaidon,
\newblock ``Learning to track with object permanence,''
\newblock in {\em CVPR}, 2021.

\bibitem{liang2022rethinking}
Chao Liang, Zhipeng Zhang, Xue Zhou, Bing Li, Shuyuan Zhu, and Weiming Hu,
\newblock ``Rethinking the competition between detection and reid in
  multiobject tracking,''
\newblock {\em TIP}, 2022.

\bibitem{chu2023transmot}
Peng Chu, Jiang Wang, Quanzeng You, Haibin Ling, and Zicheng Liu,
\newblock ``Transmot: Spatial-temporal graph transformer for multiple object
  tracking,''
\newblock in {\em WACV}, 2023.

\end{thebibliography}

\end{document}